\documentclass[journal]{IEEEtran}
\usepackage{algorithm}
\usepackage{algpseudocode}
\usepackage{algorithmicx}
\usepackage{listings}

\usepackage[utf8]{inputenc} 
\usepackage[T1]{fontenc}    
\usepackage{hyperref}       
\usepackage{url}            
\usepackage{booktabs}       
\usepackage{amsfonts}       
\usepackage{nicefrac}       
\usepackage{microtype}      
\usepackage{xcolor}         
\usepackage{threeparttable}
\usepackage{amsmath,amssymb,amsfonts}
\usepackage[pdftex]{graphicx}
\usepackage{multirow}
\usepackage{multicol}
\usepackage{adjustbox}
\usepackage{wrapfig}
\usepackage{booktabs}
\usepackage{rotating}

\ifCLASSINFOpdf
\else
\fi
\begin{document}
%
\title{Graph Generation Powered with LLMs for Boosting Multivariate Time-Series Representation Learning}
%
%
%

\author{Yucheng Wang,
        Min Wu,
        Ruibing Jin,
        Xiaoli Li,~\IEEEmembership{Fellow, IEEE,}
        Lihua Xie,~\IEEEmembership{Fellow, IEEE,}
        and Zhenghua Chen
\thanks{Yucheng Wang is with Institute for Infocomm Research, A$^*$STAR, Singapore and the School of Electrical and Electronic Engineering, Nanyang Technological University, Singapore (Email: yucheng003@e.ntu.edu.sg).}
\thanks{Min Wu and Zhenghua Chen are with Institute for Infocomm Research, A$^*$STAR, Singapore (Email: wumin@i2r.a-star.edu.sg, chen0832@e.ntu.edu.sg).}
\thanks{Ruibing Jin is with Institute for Infocomm Research, A$^*$STAR, Singapore (Email: ruibing\_jin@outlook.com).}
\thanks{Xiaoli Li is with Institute for Infocomm Research, A$^*$STAR, Singapore and the College of Computing and Data Science, Nanyang Technological University, Singapore (Email: xlli@i2r.a-star.edu.sg).}
\thanks{Lihua Xie is with the School of Electrical and Electronic Engineering, Nanyang Technological University, Singapore (Email: elhxie@ntu.edu.sg).}}

%
%

\markboth{Journal of \LaTeX\ Class Files,~Vol.~14, No.~8, August~2015}%
{Shell \MakeLowercase{\textit{et al.}}: Bare Demo of IEEEtran.cls for IEEE Journals}
%



\maketitle

\begin{abstract}
Sourced from multiple sensors and organized chronologically, Multivariate Time-Series (MTS) data involves crucial spatial-temporal dependencies. To capture these dependencies, Graph Neural Networks (GNNs) have emerged as powerful tools. As explicit graphs are not inherent to MTS data, graph generation becomes a critical first step in adapting GNNs to this domain. However, existing approaches often rely solely on the data itself for MTS graph generation, leaving them vulnerable to biases from small training datasets. This limitation hampers their ability to construct effective graphs, undermining the accurate modeling of underlying dependencies in MTS data and reducing GNN performance in this field. To address this challenge, we propose a novel framework, K-Link, leveraging the extensive universal knowledge encoded in Large Language Models (LLMs) to reduce biases for powered MTS graph generation. To harness the knowledge within LLMs, such as physical principles, we design and extract a \textit{Knowledge-Link graph} that captures universal knowledge of sensors and their linkage. To empower MTS graph generation with the knowledge-link graph, we further introduce a graph alignment module that transfers universal knowledge from the knowledge-link graph to the graph generated from MTS data. This enhances the MTS graph quality, ensuring effective representation learning for MTS data. Extensive experiments demonstrate the efficacy of K-Link for superior performance on various MTS tasks.

\end{abstract}

\begin{IEEEkeywords}
Graph Neural Network, Graph Generation, Multivariate Time-Series Data, Large Language Models
\end{IEEEkeywords}

%
\IEEEpeerreviewmaketitle

\section{Introduction}
Multivariate Time-Series (MTS) data, sourced from multiple sensors and organized chronologically, involves crucial spatial-temporal dependencies like spatial correlations among sensors and temporal patterns \cite{wang2023local,deng2021graph}. To achieve optimal performance on MTS tasks \cite{gupta2020approaches, yang2022deep,younis2023flames2graph}, it is important to learn decent representations by recognizing these dependencies. Recently, Graph Neural Networks (GNNs) have emerged as promising solutions \cite{jia2020graphsleepnet,sabbaqi2023graph,jin2024survey}, effectively capturing both spatial and temporal dependencies for superior MTS data representation compared to conventional models \cite{franceschi2019unsupervised,8437249,LIU2020113082,9484796}.

\begin{figure*}[t]
    \centering
    \includegraphics[width = 1.\linewidth]{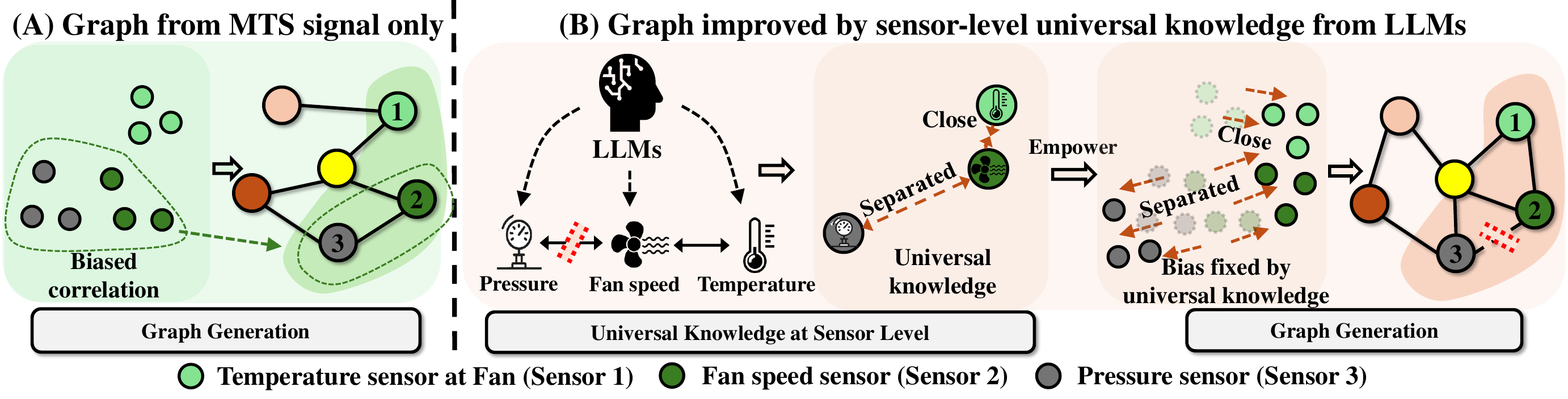}
    \caption{Graph generation to capture sensor correlations for MTS data. (A): Graph from MTS data alone is biased by incorrect sensor correlations. Sensor features of fan speed are closer to pressure than temperature, resulting in biased edge connections. (B): LLMs can encode the physical principle that fan speed correlates with temperature, not pressure. This ensures that in the feature space, fan speed should be close with temperature while being separated from pressure. By incorporating this knowledge, sensor feature distributions of MTS data can be enhanced with effective sensor correlations, leading to improved MTS graph with correct edges.}
    \label{fig:prompt_nece}
\end{figure*}

As MTS data lacks explicit graphs, graph generation is essential to adapt GNNs for MTS data. Conventional methods mainly rely on data itself for MTS graph generation to model dependencies among sensors \cite{wang2023fully,jia2020graphsleepnet}. Typically, sensor features are learned from data, with close feature distributions between sensors indicating high correlations, based on which edges are connected for MTS graph generation. However, this scheme can be easily biased by the distribution of small training datasets, leading to biased sensor correlations and limited generalizability. Using Fig. \ref{fig:prompt_nece} (A) as examples, biases may lead to sensor features of fan speed being closer to pressure than temperature, contrary to the physical rule that fan speed should be more correlated with temperature than pressure. Such biases in sensor correlations lead to incorrect edge connections, affecting MTS graph quality and thus impacting the effectiveness of GNN-based MTS representation learning.

To reduce data biases in graph generation, we propose leveraging Large Language Models (LLMs). Trained on extensive real-world data with numerous parameters \cite{xue2023promptcast,xiong2024large,yang2023harnessing}, LLMs encode comprehensive universal knowledge for sensors and their correlations, providing an effective solution to reduce biases arising from small training datasets. Using Fig. \ref{fig:prompt_nece} (B) as examples, LLMs can recognize that fan speed correlates with temperature, not pressure. This ensures that in the feature space, fan speed should be close with temperature while being separated from pressure, reflecting correct physical principles. By incorporating the knowledge of sensors and the underlying principles that link the knowledge, we can improve sensor-level feature distributions of MTS data with effective sensor correlations, enhancing MTS graph generation with correct edge connections and, subsequently, improving representation learning with GNNs for MTS data.

Adapting LLMs to empower graph generation involves two key challenges: 1. The universal knowledge of LLMs is implicitly embedded in extensive parameters. To leverage the knowledge for MTS graph generation, it is crucial to explicitly extract the knowledge, including sensor knowledge and their interconnections. 2. Effectively leveraging the knowledge-link graph to guide the learning of sensor features and their correlations for enhanced graph generation poses another significant challenge. To address these challenges, we design a novel framework, K-Link, incorporating a \textit{Knowledge-Link graph} to empower MTS graph generation for enhanced MTS representation learning with GNNs. To address the first challenge, we introduce a knowledge-link branch that extracts a knowledge-link graph from LLMs, explicitly representing universal sensor knowledge and the linkage of the knowledge. Here, sensor-level prompts are designed as queries to extract sensor knowledge from LLMs, which are then linked based on their semantic relationships. To address the second challenge, we design a graph alignment module, aiming to transfer the universal knowledge from the knowledge-link graph to the graph generated from MTS data for improved graph quality. This module includes node and edge alignment to align sensors and their relationships between two graphs, enabling comprehensive universal knowledge transfer. Through K-Link, we enhance the MTS graph quality, thus improving representation learning with GNNs for MTS data.

Our contributions are summarized as follows: 
\begin{itemize}
    \item Design and extract a knowledge-link graph from LLMs to explicitly represent universal sensor knowledge and the linkage of the knowledge, providing guidance to reduce the impact of biased sensor correlations arising from small training datasets.
    \item Design a graph alignment module to transfer universal knowledge from the knowledge-link graph to the graph generated from MTS data, thus empowering MTS graph generation and enhancing the representation learning capabilities of GNN for MTS data.
    \item Conduct extensive experiments across various MTS downstream tasks, verifying the effectiveness of K-Link for enhanced MTS data representation learning.
\end{itemize}

\section{Related Work}

\subsection{MTS Representation Learning with GNNs}
To learn representations for MTS data, traditional methods predominantly emphasized temporal dependencies using temporal encoders such as convolutional neural networks \cite{franceschi2019unsupervised,wang2023multivariate,zhang2020tapnet,eldele2021time,eldele2023self}, LSTM \cite{DU2020269,9484796,ma2020adversarial}, and Transformers \cite{wu2021autoformer,zhou2021informer}. However, these approaches often overlooked crucial spatial dependencies among sensors, limiting their representation learning capabilities for MTS data. To address this, GNNs, capable of capturing both spatial-temporal dependencies, have emerged as more effective solutions than traditional methods \cite{jin2024survey}. As MTS data inherently lacks explicit graphs, graph generation is required before GNNs. Typically, sensor features are learned from MTS data by employing temporal encoders, where close feature distributions between sensors indicate high correlations, guiding edge connections for graph generation \cite{wang2023multivariate,jia2020graphsleepnet,LI2021107878,wang2023fully}. 

While effective, existing methods mainly rely on MTS data itself for graph generation, which can be biased by the distributions of small training datasets. This limitation hampers effective generalization, impacting graph quality and thus hindering representation learning with GNNs for MTS data.
While some domain-specific approaches have incorporated external knowledge to enhance graph generation, these methods are typically tailored to specific domains, such as industrial maintenance \cite{kong2022spatio} and healthcare \cite{jia2021multi}. Consequently, they require domain-specific expertise, limiting their generalizability and broader applicability. To overcome these limitations, we propose K-Link, a method that leverages LLMs to automatically extract universal knowledge about sensors and their relationships, enhancing the graph generation process without relying on domain-specific expertise and thereby improving applicability for representation learning with GNNs for MTS data.

\subsection{LLMs for Time-Series Data}
With the capability to encode extensive knowledge through numerous parameters, LLMs can encapsulate universal sensor knowledge and underlying relationships of the knowledge, offering a promising solution to empower MTS graph generation by reducing biases.

Recently, researchers have begun exploring the potential of LLMs for time-series data analysis \cite{xue2023promptcast, zhou2023one, sun2023test, liang2024foundation}. PromptCast \cite{xue2023promptcast} firstly introduced LLMs into time-series forecasting tasks by translating numerical signals into prompts. TIME-LLM \cite{jin2024time} reimagined LLMs for general time-series forecasting by reformulating time-series to align with LLMs' capabilities. LLM4TS \cite{chang2023llm4ts} proposed a two-stage fine-tuning approach tailored for time-series forecasting, aligning LLMs with time-series data for time-series representations. These pioneering works have paved the way for further advancements in leveraging LLMs for time-series data analysis, inspiring subsequent research \cite{gruver2024large, tan2024language}. While these efforts have significantly advanced the application of LLMs to time-series tasks, they struggle to address the challenges of adapting LLMs to empower MTS graph generation. Existing efforts predominantly focus on fine-tuning LLMs without being inherently tailored for graph structures within MTS data, which poses challenges in explicitly extracting sensor knowledge and establishing relationships, thus failing to enhance MTS graph generation. Although some works, such as UrbanGPT \cite{li2024urbangpt} and ST-LLM \cite{liu2024spatial}, have proposed using LLMs for capturing spatial-temporal information in graphs, they are primarily designed for scenarios where graphs are already available, such as traffic networks. While they excel at utilizing spatial and temporal patterns within predefined graphs, they do not address the challenge of generating graphs in cases where explicit graphs are absent. For MTS tasks where graphs are not readily available—such as sensor networks deployed in machines or human bodies—the potential of utilizing LLMs to enhance graph generation remains under-explored. Accurately modeling the correlations among sensors and generating meaningful graph structures with the power of LLMs for downstream tasks represents a critical yet unmet need in this field, and this gap motivates the design of K-Link.

\section{Methodology}
\subsection{Problem Definition}

Given a training dataset $\mathcal{D}$ with $n$ labeled MTS samples, $\{X_a, y_a\}_{a=1}^n$, where each sample $X_a\in\mathbb{R}^{N\times{L}}$ with its label $y_a$ includes $N$ sensors across $L$ timestamps, our aim is to learn an encoder from this dataset, and the learned encoder can then be applied to infer the representation of a testing sample. As spatial-temporal dependencies play a critical role in MTS data, it is essential to model these dependencies through graph generation. The resulting graphs can then be processed by GNNs, enabling the learning of effective representations for downstream tasks.

Currently, existing approaches primarily employ two paradigms for graph generation: the single graph, where a single graph $G^S_a$ is created for each sample \cite{jia2020graphsleepnet}, and sequential graphs, which involve constructing $T$ sequential graphs $\{G^S_{a,t}\}_t^T$ \cite{wang2023fully}. Notably, the single graph paradigm represents a special case of the sequential graphs when $T=1$. As such, we focus on sequential graphs, allowing the framework to be readily adapted for cases involving a single graph for each sample. Here, each sequential graph within $\{G^S_{a,t}\}_t^T$ is defined as $G^S_{a,t} = \{V^S_{a,t}, E^S_{a,t}\}$, where the nodes $V^S_{a,t} = \{v_{a,t,i}^S\}_i^N$ represent sensors with features $Z^S_{a,t} = \{z_{a,t,i}^S\}_i^N$, and edges $E^S_{a,t} = \{e_{a,t,ij}^S\}_{i,j}^N$ capture sensor correlations. 

To learn effective features, these sequential graphs are processed by GNNs to capture the dependencies. Notably, the quality of graph generation directly affects the GNN’s representation learning capabilities. However, the graphs $\{G^S_{a,t}\}_t^T$ generated from MTS data can be biased by the distribution of $\mathcal{D}$. To reduce the bias, we extract the knowledge-link graph from LLMs to represent the universal sensor knowledge and the linkage of the knowledge. This knowledge-link graph aims to enhance graph generation for MTS data, supporting more robust and effective representation learning with GNNs. For simplicity, subscript $a$ is omitted in the following sections.

\begin{figure*}[htbp]
    \centering
    \includegraphics[width = 1.\linewidth]{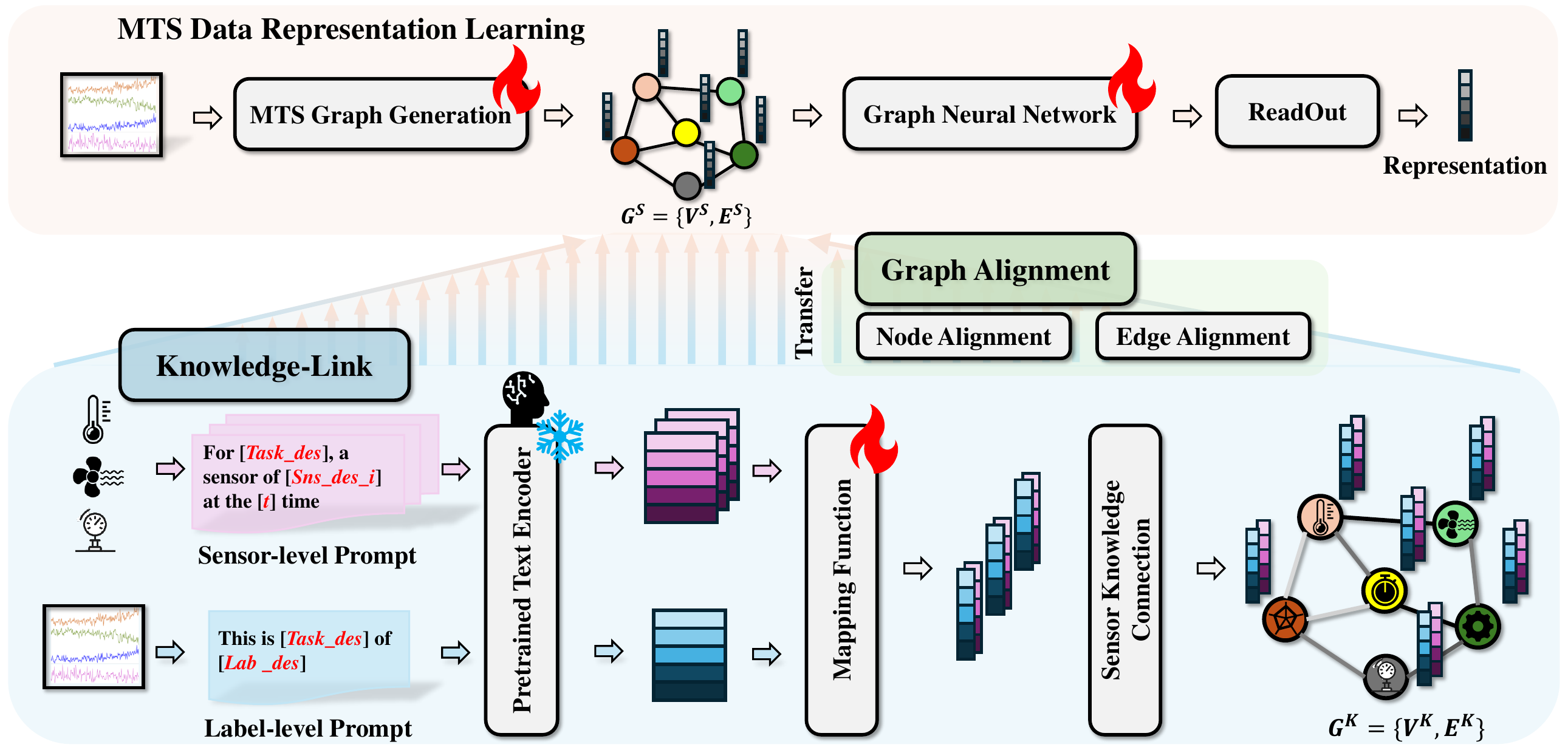}
    \caption{The overall framework, starting with an MTS graph generation and GNN branch for learning representations. In the knowledge-link branch, a knowledge-link graph is extracted from LLMs to represent universal knowledge. To unlock LLMs' potential, we design sensor-level prompts to extract sensor knowledge and label-level prompts to further enhance the sensor knowledge by considering category information. The knowledge-link graph is then defined with sensor knowledge as nodes and their semantic relationships as edges. To leverage this universal knowledge, a graph alignment module—comprising node and edge alignment—is introduced, facilitating the comprehensive transfer of knowledge from the knowledge-link graph to the MTS-generated graph, thereby enhancing MTS graph generation.}
    \label{fig:overall}
\end{figure*}
\subsection{Overall Structure}
To empower MTS graph generation and representation learning with GNNs, we propose K-Link, which incorporates universal knowledge from LLMs. As shown in Fig. \ref{fig:overall}, we start with an MTS graph generation and GNN branch for learning representations, following the structure in previous studies \cite{wang2023fully} (details in the appendix). To improve MTS graph generation, we introduce the knowledge-link module, which extracts a knowledge-link graph from LLMs to explicitly capture universal knowledge of sensors and their relationships. Recognizing the success of prompts in unlocking LLMs' potential \cite{radford2021learning}, we design prompts as queries to extract relevant sensor-level knowledge. Specifically, we develop two types of prompts: sensor-level prompts to extract universal sensor knowledge and label-level prompts to account for variations in sensor knowledge across different categories. The extracted sensor knowledge is used to construct the knowledge-link graph by establishing edges based on their semantic relationships. With the knowledge-link graph that captures universal knowledge of sensors and their relationships, we introduce a graph alignment module, comprising node and edge alignment, to comprehensively transfer the knowledge within the knowledge-link graph to the graph generated from MTS data, thus improving MTS graph generation.

\subsection{Knowledge-link Branch}
Equipped with numerous trainable parameters and trained on extensive real-world knowledge, LLMs store universal knowledge that can enhance MTS graph generation by reducing biases from small training datasets. For instance, knowledge about physical principles serves as guidance for sensor relationships. As the knowledge is implicitly embedded in LLMs' parameters, we extract a knowledge-link graph, aiming to represent universal sensor knowledge and the linkage of the knowledge.
To extract an effective knowledge-link graph from LLMs, three points need to be required:
\begin{itemize}
    \item Adhere to the similar topological structure of the MTS graph, comprising nodes and edges;
    \item Derived from the universal knowledge embedded within LLMs;
    \item Capable of representing universal sensor knowledge and linking the knowledge.
\end{itemize}
By addressing point 1, the knowledge within the knowledge-link graph can be precisely matched with MTS-generated graphs, enabling seamless and effective knowledge transfer. To achieve this, we define sequential knowledge-link graphs $\{G_t^K\}_t^T$ that align with sequential MTS graphs, where each sequential knowledge-link graph $G_t^K$ contains $V_t^K$ and $E_t^K$ to represent nodes and edges, respectively. To extract effective knowledge-link graphs by defining these nodes and edges, we leverage the universal knowledge embedded in LLMs, addressing point 2. Specifically, prompts are utilized as queries to retrieve relevant knowledge from LLMs \cite{radford2021learning,xue2023promptcast,jin2024time}. Sensor and label-level prompts are designed to extract universal sensor knowledge from LLMs, defining graph nodes. Edges are then determined by the semantic relationships of the sensor knowledge. By doing so, we can meet point 3 where nodes and edges represent universal knowledge of sensors and their linkages. We introduce details in the following parts. 

\begin{figure}[htbp]
    \centering
    \includegraphics[width = 1.\linewidth]{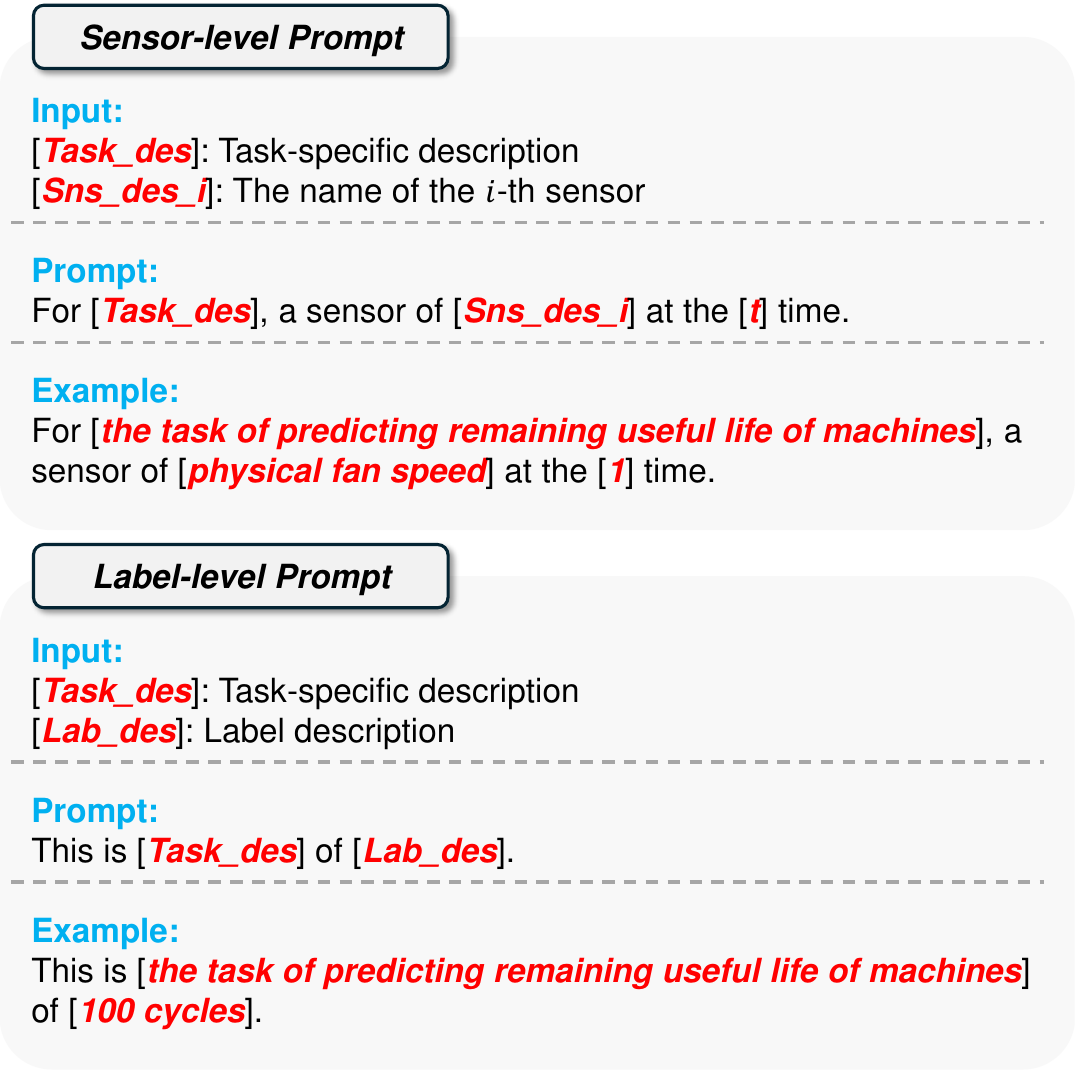}
    \caption{Prompt description.}
    \label{fig:prompt}
    \vspace{-0.2cm}
\end{figure}
\textbf{Sensor-level Prompts:} 
For nodes, we propose sensor-level prompts to derive sensor knowledge from LLMs. Given that a sensor operates within a specific scenario with defined functions, the prompts should include both contextual and sensor-specific information to access the relevant sensor knowledge. The former outlines the scenarios in which the sensor is deployed, while the latter provides the detailed functions of the sensor itself. To meet these requirements, we design prompts as shown in Fig. \ref{fig:prompt}, `For [$Task\_des$], a sensor of [$Sns\_des\_i$]', where [$Task\_des$] provides task-specific context, e.g., `For [\textit{the task of remaining useful life prediction of machines}]', and [$Sns\_des\_i$] represents the name of sensor $i$, e.g., `a sensor of [\textit{physical fan speed}]'. Additionally, considering the need for empowering sequential graphs, we extend the sensor-level prompts by incorporating temporal information $t$, so extended prompts are: $p_{i,T} =$ `For [$Task\_des$], a sensor of [$Sns\_des\_i$] at the [$t$] time', where $t$ represents the $t$-th sequential graph. By applying the prompts for all sensors, we obtain $\{\{p_{t,i}\}_i^N\}_t^{T}$ for each sample. Notably, as $t$ is automatically generated based on the data, the creation of the prompt only requires descriptions for the task itself and the sensors, which are inherently available in real-world systems.

\textbf{Label-level Prompts:} Recognizing that samples in different categories may exhibit distinct trends in sensor relations, we design label-level prompts. For example, in predicting the remaining useful life of a machine, temperature may correlate more strongly with fan speed during the degradation stage than during the health stage, given that a degrading fan generates more heat due to increased friction. To account for such category-specific information, we introduce label-level prompts to complement sensor-level prompts. The label-level prompt is defined for each sample, as shown in Fig. \ref{fig:prompt}, designed as $P=$ `This is [$Task\_des$] of [$Lab\_des$]', where [$Lab\_des$] is the sample label's description, and [$Task\_des$] provides contextual information of the task. With the incorporation of the label-level prompt, we can enhance the sensor-level prompts to capture category-specific sensor relations.

\textbf{Nodes:} 
Utilizing the sensor and label-level prompts, we extract knowledge from LLMs to generate nodes for the knowledge-link graph. In this work, we employ GPT-2 as the LLM due to its exceptional efficiency and effectiveness in encapsulating rich universal knowledge \cite{radford2021learning}. Trained on vast amounts of real-world data, this LLM captures a comprehensive understanding of sensors. By leveraging its pretrained text encoder $\mathcal{F}^K$, we can extract semantic features from prompts $z^K_{t,i} = \text{concat}(z^p_{t,i},g^p)$, where $z^p_{t,i}=\mathcal{F}^K(p_{t,i})$ and $g^p=\mathcal{F}^K(P)$, combining two levels of prompts to generate comprehensive universal sensor knowledge for the nodes in the knowledge-link graph. Finally, we obtain the nodes of the knowledge-link graph as $V_t^K = \{v^K_{t,i}\}_i^{N}$ with features $Z_t^K = \{z^K_{t,i}\}_i^{N}$.

\textbf{Edges:} 
The edges of the knowledge-link graph should be determined by the semantic relationships of sensor knowledge, with highly correlated sensors represented by strong connections in the knowledge-link graph. For example, the temperature of a fan is highly correlated with the fan's speed, suggesting a strong connection between these variables in the graph. To quantify these relationships, we propose utilizing the semantic features of sensor knowledge encoded by LLMs' text encoder. These features present sensor semantic information, so their similarities can effectively represent semantic relationships, i.e., large semantic similarities indicate strong links of the knowledge, and vice versa. Here, we adopt dot-product for similarity computation $e^K_{t,ij} = z^K_{t,i}(z^K_{t,j})^T$. With this approach applied across all sequential graphs, we obtain the sequential knowledge-link graph $\{G_t^K\}_t^T$, where $G_t^K = (V_t^K, E_t^K)$ is for the $t$-th graph. Here, $V_t^K = \{v^K_{t,i}\}_i^{N}$ represents sensor knowledge, with features $Z_t^K = \{z^K_{t,i}\}_i^{N}$ signifying semantic features of the knowledge. $E_t^K = \{e^K_{t,ij}\}_{i,j=1}^{N}$ represents edges, denoting the link strength of sensor knowledge. The knowledge-link graph can then empower the MTS graph generation.

\begin{figure*}[htbp]
    \centering
    \includegraphics[width = 1.\linewidth]{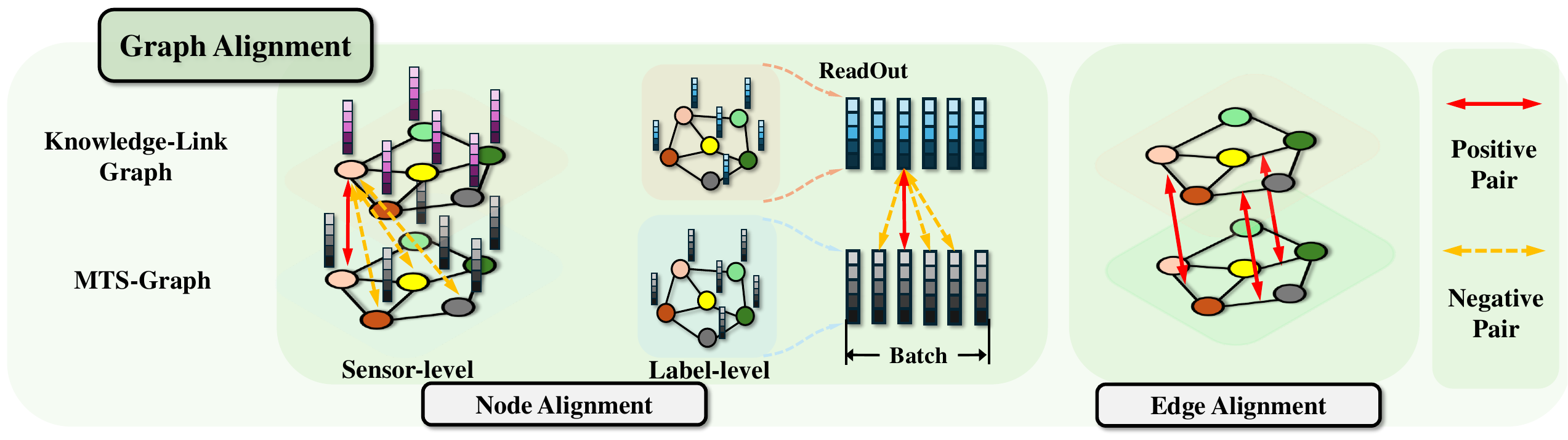}
    \caption{The graph alignment, including node and edge alignment, to comprehensively transfer the universal knowledge from the knowledge-link graph to the MTS graph. Node alignment is further divided into sensor-level and label-level alignment to ensure a balanced and effective transfer of the sensor knowledge at both levels.}
    \label{fig:overall}
\end{figure*}
\subsection{Graph Alignment}
The knowledge-link graph, enriched with extensive universal knowledge extracted from LLMs, serves as a powerful tool to enhance MTS graph quality by mitigating biases arising from small training datasets. To transfer the universal knowledge from the knowledge-link graph to the MTS graph, we propose a graph alignment module, which includes node and edge alignment to achieve comprehensive knowledge transfer, as shown in Fig. \ref{fig:overall}.

\textbf{Node Alignment:}
The nodes' semantic features of the knowledge-link graph originate from two sources: sensor-level and label-level prompts, offering universal knowledge of sensors and categories, respectively. While directly aligning the entire features of each node between two graphs is straightforward, it may pose challenges in achieving a balanced transfer of sensor-level and label-level knowledge. To address this, we propose to separate node alignment into sensor-level and label-level alignment. This approach allows for a better balance between the two levels, ensuring the effective transfer of universal knowledge within the knowledge-link graph.

Sensor-level alignment is achieved within each MTS sample. We expect to align corresponding sensors across two graphs, distinguishing them from other sensors. This alignment ensures that features of each sensor learned from data can be precisely matched with the sensor's universal knowledge. To achieve this, we employ the InfoNCE loss of contrastive learning \cite{radford2021learning} for the sensor-level alignment. The sensor-level alignment, defined in Eq. (\ref{eq:nodects}), is first conducted within each sequential graph, and the contrastive loss is then aggregated across all sequential graphs. In this formulation, $z_{t,i}^S$ and $z_{t,i}^p$ represent the features from MTS data and the sensor-level knowledge, respectively, for sensor $i$ from the $t$-th graph. $\hat{\mathcal{V}}_{t,i}$ represents the set of nodes excluding node $i$ for the $t$-th graph, $f_{sim}(\cdot,\cdot)$ measures similarity implemented by the dot product, and $\tau$ is a temperature parameter.
\begin{equation}
    \label{eq:nodects}
        \mathcal{L}_{S} = -\frac{1}{N}\sum_t^{T}\sum_i^{N}log\frac{exp(f_{sim}(z_{t,i}^S, z_{t,i}^p)/\tau)}{\sum_{m\in\hat{\mathcal{V}}_{t,i}}exp(f_{sim}(z_{t,i}^S, z_{t,m}^p)/\tau)}.
\end{equation}
Label-level alignment is carried out within each training batch, as the label-level prompt is specified for each sample. We expect to align corresponding samples across two sides, distinguishing them from other samples. To obtain the features for each sample, we employ a readout function by stacking all sensor' features as $g^S = \text{concat}(z_{1,1}^S,...,z_{T,N}^S)$. Then, the InfoNCE loss is utilized to achieve the label-level alignment, as shown in Eq. (\ref{eq:graphcts}), where $g^{S}_a$ and $g^{p}_a$ represent the global features from sample $a$ and its label-level knowledge, respectively. $\hat{\mathcal{U}}_a$ denotes the set of samples excluding $a$, and $B$ is the batch size. Notably, the label-level alignment is performed exclusively during the training stage, ensuring that no label information is leaked during the inference stage.
\begin{equation}
    \label{eq:graphcts}
        \mathcal{L}_{L} = -\frac{1}{B}\sum_{a=1}^Blog\frac{exp(f_{sim}(g^S_a,g^p_a)/\tau)}{\sum_{u\in\hat{\mathcal{U}}_{a}}exp(f_{sim}(g^S_a,g^p_u)/\tau)}.
\end{equation}
\textbf{Edge Alignment:}
Edge alignment aims to transfer semantic relationships of sensor knowledge into sensor correlations learned from MTS data. To achieve this, we propose aligning each edge between two graphs, as each edge represents the correlation between two sensors. This alignment minimizes the Mean Square Error (MSE), as shown in Eq. (\ref{eq:edgects}), ensuring that the sensor correlations from MTS data are consistent with the semantic sensor relationships encoded in the knowledge-link graph. 
This enables the mitigation of the biased sensor correlations from small training datasets to enhance the MTS graph.
\begin{equation}
    \label{eq:edgects}
        \mathcal{L}_{E} = \frac{1}{2}\sum_t^{T}\sum_{i}^{N}\sum_j^{N}(e_{t,ij}^S - e_{t,ij}^K)^2/N^2.
\end{equation}
In Eq. (\ref{eq:overaloss}), we combine the node and edge alignment together to empower MTS graph generation with the knowledge-link graph. Notably, both sensor-level and edge alignment are performed within each sample, denoted as $\mathcal{L}_{a,S}$ and $\mathcal{L}_{a,E}$ for the $a$-th sample. $\lambda_{S}$, $\lambda_{L}$, and $\lambda_{E}$ are hyperparameters to balance these losses. $\mathcal{L}_D$ is the loss determined by downstream tasks. 
\begin{equation}
    \label{eq:overaloss}
    \mathcal{L} = \mathcal{L}_D+\lambda_{S}\sum_a^B\mathcal{L}_{a,S} + \lambda_{L}\mathcal{L}_{L} + \lambda_{E}\sum_a^B\mathcal{L}_{a,E}.
\end{equation}
With the graph alignment module, we can effectively transfer the knowledge embedded within the knowledge-link graph, ensuring that the universal knowledge of sensors and their linkage guides the graph generation process. By doing so, the framework enhances the quality of the graphs generated from MTS data, thereby improving GNN-based representation learning. The pseudocode of K-Link can be found in Algorithm \ref{alg:overall}, which outlines the detailed steps involved in the framework.

\begin{algorithm}[!h]
\caption{Pseudocode of K-Link.}
\label{alg:overall}
\definecolor{codeblue}{rgb}{0.25,0.5,0.5}
\definecolor{codekeyword}{rgb}{0.13,0.13,1} 
\lstset{
  backgroundcolor=\color{white},
  basicstyle=\fontsize{7.2pt}{7.2pt}\ttfamily\selectfont,
  columns=fullflexible,
  breaklines=true,
  captionpos=b,
  commentstyle=\fontsize{7.2pt}{7.2pt}\color{codeblue},
  keywordstyle=\fontsize{7.2pt}{7.2pt}\color{codekeyword}, 
  language=Python,
  morekeywords={def, in}, 
}
\begin{lstlisting}[language=python]
# X, MTS sample with [N,L], N: number of sensors, L: time length
# y, label of X
# f, patch size
# L_hat, number of patches for sequential graphs
# ts_enc, time series encoders, such as CNN or LSTM

# sns_prompt, prompts for N sensors across L_hat times
# lab_prompt, prompt for the label of X
# F_text, pretained text encoder from LLMs, e.g., GPT-2
# F_M, mapping function with [512, d_h]
# F_l, function to map the features from GNN for downstream tasks

train()

# Step 1: graph generation from MTS data
sig_patch = sample_partition(X, f) # [L_hat,N,f] 
sig_node_feat = ts_enc(sig_patch) # [L_hat,N,d_h]
sig_edge = sig_edge_comp(sig_node_feat) # [L_hat,N,N]

# Step 2: extract knowledge-link graph from LLMs
sns_prompt_feat = F_M(F_text(sns_prompt)) # [L_hat,N,d_h]
lab_prompt_feat = F_M(F_text(lab_prompt)) # [d_h]
kl_node_feat = concat(sns_prompt_feat,exp_dim(lab_prompt_feat)) # [L_hat,N,2*d_h]
kl_edge = kl_edge_comp(kl_node_feat) # [L_hat,N,N]

# Step 3: compute losses for graph alignment
loss_sns = sns_level_align(sig_node_feat, sns_prompt_feat)
loss_lab = lab_level_align(readout(sig_node_feat), lab_prompt_feat)
loss_edge = edge_align(sig_edge,kl_edge)

# Step 4: compute downstream task loss
loss_D = loss_comp(F_l(MPNN(sig_node_feat,sig_edge)),y)

# Step 5: Derive overall loss and update model parameters
loss = loss_D + lambda_s * loss_sns + lambda_L * loss_lab + lambda_e * loss_edge

loss.backward()
optim.step()

\end{lstlisting}
\end{algorithm}

\section{Experimental Results}
\subsection{Experimental Setting and Dataset Details}
\subsubsection{Dataset Details}


Inspired by previous works \cite{wang2024sea++, wang2023fully}, we evaluate K-Link on multiple MTS downstream datasets, encompassing both regression and classification tasks. These evaluations assess the model's ability to learn effective representations for predicting continuous and discrete labels.

\textbf{Regression tasks:} We utilize the FD002 and FD004 datasets from C-MAPSS \cite{saxena2008CMAPSS}, both of which focus on predicting a continuous value representing the Remaining Useful Life (RUL) of a machine before failure. As K-Link relies on label-level prompts to enhance sensor knowledge, the labels must provide categorical information to capture variations across different categories. RUL values effectively serve this purpose by offering category insights, such as the degradation stages of a machine, making these datasets well-suited for evaluating K-Link's performance. Furthermore, the continuous nature of RUL values also provides a strong benchmark for evaluating a model's ability to predict continuous labels based on learned representations. Meanwhile, the datasets include samples collected under diverse fault modes and operating conditions, representing complex, real-world scenarios \cite{chen2020machine} and enabling a robust assessment of a model's practical applicability. The datasets consist of measurements from 21 sensors used for machine status detection. In line with previous studies \cite{liu2020remaining, wang2023sensor}, we limit the analysis to 14 sensors, as the remaining sensors exhibit constant values and contribute no additional information. As the samples cover the entire lifecycle of the machines, we apply sliding-window-based pre-processing to extract relevant samples. To further refine the labels, we use piecewise linear RUL estimation \cite{al2019hybrid, xu2022sgbrt}, which ensures smooth and realistic predictions of machine degradation. For consistency with prior studies \cite{wang2023multivariate, wang2023fully}, we adopt the standardized train-test splits provided in the dataset for model training and evaluation.

\textbf{Classification tasks:} We utilize the UCI-HAR \cite{anguita2012human}, ISRUC \cite{khalighi2016isruc}, and WISDM \cite{kwapisz2011activity} datasets, alongside CharacterTrajectories, SelfRegulationSCP1, and FingerMovements from \cite{bagnall2018uea}, to evaluate the model's ability to predict discrete labels across diverse classification tasks. We followed the previous study \cite{wang2023graph} to process all these datasets. In particular, the UCI-HAR and WISDM datasets focus on human activity recognition. UCI-HAR contains data from nine channels corresponding to various sensors, such as accelerometers and gyroscopes, attached to the human body. The dataset includes records of six activities: walking, walking upstairs, walking downstairs, standing, sitting, and lying down. Similarly, the WISDM dataset, derived from three channels with accelerometer sensors, captures the same six activities as UCI-HAR. ISRUC was collected for sleep stage classification and comprises nine channels recording data across five sleep stages: wakefulness, N1, N2, N3, and REM. Due to its time length of 3000, which can be computationally expensive for training, we downsample the data at intervals of ten, reducing the time length to 300. For HAR, WISDM, and ISRUC, we randomly split the datasets into 60\% for training, 20\% for validation, and 20\% for testing. CharacterTrajectories captures the process of writing characters using three channels to record the coordinates and pen forces for 26 alphabet letters. SelfRegulationSCP1 was collected from a healthy subject moving a cursor vertically on a computer screen, with six channels recording cortical potentials. FingerMovements aims to classify left-hand or right-hand finger movements using data from 28 channels. For these three datasets, we adopt the predefined train-test splits provided by the datasets.

\subsubsection{Experimental Setting}

Our experiments were conducted on an NVIDIA GeForce RTX 3080Ti GPU using PyTorch 1.9. The model was trained with the ADAM optimizer over 50 epochs. During the training phase, an LLM was integrated to enhance graph generation, and it was removed during the inference phase, where only the enhanced GNN-based framework, including MTS graph generation and GNN, was used for testing. Following prior work \cite{radford2021learning}, we employed GPT-2 as the LLM due to its exceptional efficiency and effectiveness in encapsulating rich universal knowledge. For the GNN-based framework, we adopted the architecture proposed in \cite{wang2023fully}, which uses a fully-connected graph constructed from segmented patches, and a combination of CNNs and message-passing neural networks was employed to capture temporal and spatial dependencies, respectively. The implementation details can be found in the appendix. Additionally, discussions regarding the selection of alternative LLMs and backbone architectures are provided in Section \ref{sec:discuss}.

To evaluate regression tasks, we utilized Root Mean Square Error (RMSE) and the Score function \cite{chen2020machine,wang2023sensor}. The RMSE measures the overall prediction error, treating early and late predictions equally, while the Score function imposes a higher penalty on late predictions, as they typically result in more significant losses in real-world applications.  Thus, the Score function is often considered more critical than RMSE due to its alignment with real-world production priorities. The formulas for these metrics are as follows:
\begin{equation}
    \label{eq12} \text{RMSE} = \sqrt{\frac{\sum_{a=1}^n(\hat{y}_a - y_a)^2}{n}},
\end{equation}
\begin{equation}
\setlength{\jot}{3pt}
    \begin{array}{l}
    Score_a=
  \left\{\begin{matrix} 
  e^{-\frac{\hat{y}_a - y_a}{13}}-1, & \textit{if } {\hat{y}_a < y_a}, \\ 
  e^{\frac{\hat{y}_a - y_a}{10}} - 1, & \textit{otherwise},
\end{matrix}\right.
\end{array} 
\nonumber
\end{equation}
\begin{equation}
\label{eq13} 
    Score = \sum_{a=1}^nScore_a.
\end{equation}
Here, $\hat{y}_a$ represents the predicted RUL, $y_a$ the actual RUL, and $n$ the total number of samples in the dataset. \textit{Lower RMSE and Score values indicate better performance.}

For classification tasks, we adopted standard Accuracy (Acc.) and Macro-averaged F1-score (MF1) \cite{eldele2021time,meng2022mhccl} as evaluation metrics, \textit{with higher values indicating superior performance.} To mitigate randomness, each experiment was repeated ten times, with the average results reported alongside standard deviations to demonstrate model robustness.

\subsection{Comparisons with SOTAs}
\begin{table}[htbp]
\scriptsize
  \centering
  \caption{Comparisons with SOTA approaches for regression tasks.}
  \begin{threeparttable}
    \begin{tabular}{lcccc}
    \toprule
    \toprule
    \multirow{2}[2]{*}{Variants} & \multicolumn{2}{c}{\textbf{FD002}} & \multicolumn{2}{c}{\textbf{FD004}} \\
          & RMSE \textbf{$\downarrow$}  & Score \textbf{$\downarrow$} & RMSE \textbf{$\downarrow$} & Score \textbf{$\downarrow$}\\
    \midrule
    InFormer & 13.20±0.15 & 715±71 & 14.16±0.49 & 1023±201 \\
    AutoFormer & 16.51±0.47 & 1248±112 & 20.31±0.14 & 2291±122 \\
    TS-TCC & 15.08±0.49 & 931±84 & 16.62±0.49 & 1147±134 \\
    TS2Vec & 15.67±0.51 & 1047±192 & 16.36±0.47 & 1675±157 \\
    PatchTST & 15.51±0.34 & 1143±116 & 16.87±0.32 & 1409±113 \\
    TimesNet & \underline{12.95±0.09} & 773±69 & 14.37±0.15 & 990±127 \\
    TSLANet & 14.01±0.21 & 932±89 & 14.33±0.31 & 992±96 \\
    \midrule
    HAGCN & 14.92±0.12 & 1086±87 & 14.66±0.25 & 880±150 \\
    HierCorrPool & 13.23±0.31 & 709±61 & 13.86±0.32 & 854±68 \\
    MAGNN & 13.09±0.13 & 714±57 & 14.30±0.26 & 978±137 \\
    FC-STGNN & 13.04±0.13 & 738±49 & \underline{13.62±0.25} & \underline{816±63} \\
    LOGO  & 13.01±0.15 & \underline{696±43} & 13.83±0.25 & 917±72 \\
    \midrule
    Ours  & \textbf{12.87±0.14} & \textbf{634±30} & \textbf{13.36±0.22} & \textbf{786±61} \\
    \bottomrule
    \bottomrule
    \end{tabular}%
\begin{tablenotes}
        \item \textbf{$\downarrow$}: Lower values are better
      \end{tablenotes}
\end{threeparttable}
  \label{tab:sotas_pre}%
\end{table}%

\begin{table*}[htbp]
  \centering
  \caption{Comparisons with SOTA approaches for classification tasks.}
  \begin{adjustbox}{width = 1\linewidth,center}
  \begin{threeparttable}
    \begin{tabular}{lcccccccccccc}
    \toprule
    \toprule
    \multirow{2}[2]{*}{Variants} & \multicolumn{2}{c}{\textbf{UCI-HAR}} & \multicolumn{2}{c}{\textbf{ISRUC}} & \multicolumn{2}{c}{\textbf{WISDM}} & \multicolumn{2}{c}{\textbf{CharacterTrajectories}} & \multicolumn{2}{c}{\textbf{SelfRegulationSCP1}} & \multicolumn{2}{c}{\textbf{FingerMovements}} \\
          & Acc. \textbf{$\uparrow$} & MF1 \textbf{$\uparrow$}   & Acc. \textbf{$\uparrow$} & MF1 \textbf{$\uparrow$}   & Acc. \textbf{$\uparrow$} & MF1 \textbf{$\uparrow$}   & Acc. \textbf{$\uparrow$} & MF1 \textbf{$\uparrow$}   & Acc. \textbf{$\uparrow$} & MF1 \textbf{$\uparrow$}   & Acc. \textbf{$\uparrow$} & MF1 \textbf{$\uparrow$} \\
    \midrule
    InFormer & 90.23±0.48 & 90.23±0.47 & 72.15±2.41 & 68.67±3.42 & 94.13±0.41 & 91.43±0.74 & 83.39±4.81 & 81.82±5.39 & 90.71±0.39 & 90.69±0.40 & 53.50±1.36 & 47.50±5.96 \\
    AutoFormer & 56.70±0.81 & 54.41±1.74 & 43.75±0.95 & 37.88±2.43 & 60.45±1.61 & 42.05±5.57 & 89.00±0.92 & 82.18±0.70 & 54.74±9.26 & 41.48±5.58 & 55.30±3.13 & 50.95±7.79 \\
    TS-TCC & 90.82±0.12 & 90.88±0.13 & 71.98±0.69 & 68.31±0.37 & 84.85±0.34 & 74.31±1.31 & 97.54±0.07 & 97.38±0.07 & 84.98±0.74 & 84.89±0.79 & 52.67±0.94 & 52.41±1.11 \\
    TS2Vec & 94.14±0.45 & 94.21±0.47 & 73.94±0.10 & 70.29±0.28 & 75.27±0.24 & 68.87±0.58 & \textbf{99.36±0.10} & \textbf{99.31±0.11} & 80.20±1.00 & 80.13±0.93 & 57.00±2.94 & 56.69±2.82 \\
    PatchTST & 84.21±0.24 & 84.11±0.34 & 71.23±0.67 & 69.21±2.12 & 90.45±0.32 & 88.43±0.63 & 94.64±0.18 & 94.43±0.21 & 75.69±0.35 & 75.67±0.35 & 57.81±0.52 & 57.24±0.78 \\
    TimesNet & 91.36±0.12 & 91.34±0.15 & 75.70±0.64 & 73.69±1.24 & 92.42±0.52 & 90.14±0.72 & 95.53±0.12 & 95.54±0.13 & 91.49±0.57 & 91.48±0.57 & 60.42±3.12 & 60.12±3.62 \\
    TSLANet & \underline{96.18±0.62} & \underline{96.17±0.62} & 78.86±0.39 & 75.88±1.40 & 87.17±0.33 & 82.51±0.27 & \underline{99.16±0.06} & \underline{99.12±0.06} & 82.14±1.40 & 82.01±1.43 & 55.67±2.36 & 55.42±2.54 \\
    \midrule
    HAGCN & 80.79±0.77 & 81.08±0.75 & 66.59±0.29 & 60.20±2.24 & 88.18±0.62 & 83.65±0.74 & 91.24±1.86 & 90.76±2.02 & 88.90±0.83 & 88.87±0.82 & 59.30±1.27 & 57.76±2.49 \\
    HierCorrPool & 93.81±0.26 & 93.79±0.28 & 79.31±0.60 & 76.25±0.72 & 87.00±0.18 & 81.71±0.36 & 97.25±0.06 & 97.24±0.17 & 90.47±0.41 & 90.47±0.41 & 62.10±1.92 & 60.86±2.81 \\
    MAGNN & 90.91±0.99 & 90.79±1.08 & 68.13±2.54 & 64.31±5.25 & 89.25±0.66 & 85.25±0.91 & 91.26±0.57 & 90.82±0.62 & 89.41±0.82 & 89.38±0.84 & 57.80±2.52 & 56.84±4.20 \\
    FC-STGNN & 95.81±0.24 & 95.82±0.24 & \underline{80.87±0.21} & \underline{78.79±0.55} & \underline{95.41±0.23} & \underline{93.85±0.28} & 97.36±0.27 & 97.18±0.28 & 90.93±0.72 & 90.91±0.72 & 61.10±1.42 & 60.74±1.46 \\
    LOGO  & 95.06±0.19 & 95.07±0.20 & 76.80±0.00 & 74.69±0.34 & 92.26±0.08 & 91.31±0.14 & 94.54±0.23 & 94.12±0.20 & \underline{91.04±0.37} & \underline{91.03±0.37} & \underline{64.00±2.00} & \underline{62.59±2.91} \\
    \midrule
    Ours  & \textbf{96.87±0.12} & \textbf{96.92±0.12} & \textbf{81.37±0.20} & \textbf{79.36±0.49} & \textbf{97.12±0.17} & \textbf{95.85±0.20} & 98.11±0.07 & 97.97±0.08 & \textbf{91.94±0.48} & \textbf{91.94±0.48} & \textbf{64.70±1.26} & \textbf{64.30±1.47} \\
    \bottomrule
    \bottomrule
    \end{tabular}%
\begin{tablenotes}
        \item \textbf{$\uparrow$}: Higher values are better
      \end{tablenotes}
\end{threeparttable}
    \end{adjustbox}
  \label{tab:sotas_cls}%
\end{table*}%



We benchmark K-Link against two types of State-Of-The-Art (SOTA) methods, including conventional approaches primarily focusing on temporal dependencies and GNN-based approaches. The first type of approaches include InFormer \cite{zhou2021informer}, AutoFormer \cite{wu2021autoformer}, TS-TCC \cite{eldele2021time}, TS2Vec \cite{yue2022ts2vec}, PatchTST \cite{nie2022time}, TimesNet \cite{wu2022timesnet}, and TSLANet \cite{eldele2024tslanet}. GNN-based approaches HAGCN \cite{LI2021107878}, HierCorrPool \cite{wang2023multivariate}, MAGNN \cite{chen2023multi}, FC-STGNN \cite{wang2023fully}, and LOGO \cite{wang2023local}. All the methods were implemented using their original configurations to ensure fair comparisons. Additionally, some domain-specific methods indeed enhance graph generation through domain expertise. However, these methods lack generalizability, and therefore we discussed and compared them with our K-Link separately in Section \ref{sec:discuss}.

Tables \ref{tab:sotas_pre} and \ref{tab:sotas_cls} showcase the superior performance of K-Link compared to SOTAs in regression and classification tasks, respectively. From these results, GNN methods outperform conventional approaches in most cases, emphasizing their effectiveness in capturing both spatial and temporal dependencies for better representation learning on MTS data. However, these methods are limited by training with raw data alone, which compromises the quality of the generated graphs and, consequently, the representation learning with GNNs. In contrast, K-Link achieves superior performance across tasks. For example, in regression tasks, K-Link demonstrates notable improvements of 8.91\% and 3.67\% regarding score on FD002 and FD004, respectively, compared to the second-best methods, both of which are GNN-based frameworks. Furthermore, K-Link exhibits the small standard deviation, highlighting its robust performance. Similar enhancements are observed on classification tasks, where K-Link achieves the best performance in most cases. For example, K-Link improves by 1.71\% compared to the second-best method, i.e., FC-STGNN, on WISDM regarding accuracy. 

These results underscore the effectiveness of K-Link. While existing GNN-based methods generally surpass conventional approaches, their reliance on training solely with data limits the quality of the generated MTS graphs, thereby constraining their overall performance. By leveraging knowledge-link graphs extracted from LLMs to enhance MTS graph generation, K-Link significantly improves graph quality by effectively capturing dependencies within MTS data. This ability enables K-Link to outperform SOTA methods, including advanced GNN-based frameworks.

\subsection{Ablation Study}
\begin{table}[htbp]
\scriptsize
  \centering
  \caption{The ablation study for regression tasks}
  \begin{threeparttable}
    \begin{tabular}{lcccc}
    \toprule
    \toprule
    \multirow{2}[2]{*}{Variants} & \multicolumn{2}{c}{\textbf{FD002}} & \multicolumn{2}{c}{\textbf{FD004}} \\
          & RMSE \textbf{$\downarrow$}  & Score \textbf{$\downarrow$} & RMSE \textbf{$\downarrow$}  & Score \textbf{$\downarrow$} \\
    \midrule
    w/o K-Link & 13.85$\pm$0.19 & 804$\pm$91 & 14.39$\pm$0.13 & 968$\pm$56 \\
    \midrule
    w/o node & 13.57$\pm$0.25 & 725$\pm$56 & 14.14$\pm$0.19 & 884$\pm$51 \\
    w/o node (sensor) & 13.39$\pm$0.10 & 713$\pm$56 & 14.15$\pm$0.30 & 969$\pm$73 \\
    w/o node (label) & 13.36$\pm$0.13 & 712$\pm$22 & 13.94$\pm$0.12 & 881$\pm$79 \\
    \midrule
    w/o edge & 13.48$\pm$0.05 & 742$\pm$30 & 13.62$\pm$0.09 & 896$\pm$79 \\
    \midrule
    w/ index prompt & 13.30$\pm$0.08 & 702$\pm$43 & 13.76$\pm$0.22 & 867$\pm$75 \\
    \midrule
    Ours  & \textbf{12.87$\pm$0.14} & \textbf{634$\pm$30} & \textbf{13.36$\pm$0.22} & \textbf{786$\pm$61} \\
    \bottomrule
    \bottomrule
    \end{tabular}%
\begin{tablenotes}
        \item \textbf{$\downarrow$}: Lower values are better
      \end{tablenotes}
\end{threeparttable}
  \label{tab:abla_pre}%
\end{table}%
\begin{table}[htbp]
\scriptsize
  \centering
  \caption{The ablation study for classification tasks}
  \begin{threeparttable}
    \begin{tabular}{lcccc}
    \toprule
    \toprule
    \multirow{2}[2]{*}{Variants} & \multicolumn{2}{c}{\textbf{HAR}} & \multicolumn{2}{c}{\textbf{ISRUC}} \\
          & Acc. \textbf{$\uparrow$} & MF1 \textbf{$\uparrow$}   & Acc. \textbf{$\uparrow$} & MF1 \textbf{$\uparrow$} \\
    \midrule
    w/o K-Link & 95.62$\pm$0.16 & 95.64$\pm$0.18 & 79.83$\pm$0.14 & 78.11$\pm$0.55 \\
    \midrule
    w/o node & 96.23$\pm$0.12 & 96.26$\pm$0.10 & 80.19$\pm$0.35 & 78.63$\pm$0.66 \\
    w/o node (sensor) & 96.51$\pm$0.09 & 96.56$\pm$0.10 & 80.42$\pm$0.28 & 78.81$\pm$0.49 \\
    w/o node (label) & 96.35$\pm$0.13 & 96.40$\pm$0.13 & 80.42$\pm$0.20 & 78.72$\pm$0.35 \\
    \midrule
    w/o edge & 96.15$\pm$0.12 & 96.18$\pm$0.13 & 80.10$\pm$0.19 & 78.59$\pm$0.59 \\
    \midrule
    w/ index prompt & 96.49$\pm$0.10 & 96.52$\pm$0.11 & 80.43$\pm$0.39 & 78.70$\pm$0.72 \\
    \midrule
    Ours  & \textbf{96.87$\pm$0.12} & \textbf{96.92$\pm$0.12} & \textbf{81.37$\pm$0.20} & \textbf{79.36$\pm$0.49} \\
    \bottomrule
    \bottomrule
    \end{tabular}%
\begin{tablenotes}
        \item \textbf{$\uparrow$}: Higher values are better
      \end{tablenotes}
\end{threeparttable}
  \label{tab:abla_cls}%
\end{table}%

The ablation study demonstrates the effectiveness of each improvement. The first variant, `w/o knowledge-link', removes the knowledge-link graph, utilizing the vanilla GNN encoder for representation learning. The second variant, `w/o node', eliminates the entire node alignment module. The third and fourth variants, `w/o node (sensor)' and `w/o node (label)', evaluate the effects of removing sensor-level and label-level alignment, respectively, while retaining another component for node alignment. The fifth variant, `w/o edge', explores the effect of removing edge alignment. The final variant, `w/ index prompt', employs prompts like `For [$Ta$], A sensor of [index]', replacing sensor names with numbers (1, 2, etc.), to investigate the utility of sensor knowledge introduced by sensor names.

Tables \ref{tab:abla_pre} and \ref{tab:abla_cls} present the results of the ablation study on regression and classification tasks respectively. We utilize the Score results on FD002 as an example for the following analysis. Comparing our complete method with the `w/o knowledge-link' variant, we observe a significant improvement of 21.1\%, indicating the effectiveness of the knowledge-link graph. By unlocking the power of LLMs, K-Link enhances MTS graph generation, thus learning improved representations for downstream tasks. Examining the knowledge-link graph components, both nodes and edges are crucial. Removing node and edge alignment, we observe performance drops of 12.5\% and 14.5\%, respectively, emphasizing the importance of universal sensor knowledge and their links in the knowledge-link graph. Within node alignment, the removal of sensor-level and label-level alignment leads to performance decreases of 11.1\% and 10.9\%, respectively, highlighting the significance of sensor-level and label-level prompts. Finally, replacing sensor names with index numbers results in a 9.68\% performance drop, demonstrating the importance of the knowledge within sensor names. Notably, this variant with index prompts achieves slightly better results than the variant without sensor-level alignment, indicating that the contextual information in the sensor-level prompt can contribute to performance enhancements.

The ablation study underscores the effectiveness of the knowledge-link graph, emphasizing the importance of its nodes and edges in representing universal sensor knowledge and the links of the knowledge. By incorporating the knowledge-link graph to reduce the biases arising from small training datasets, we enhance MTS graph generation, leading to improved representation learning with GNNs and better performance for downstream tasks.

\subsection{Sensitivity Analysis}

\begin{figure}[htbp!]
    \centering\includegraphics[width = .9\linewidth]{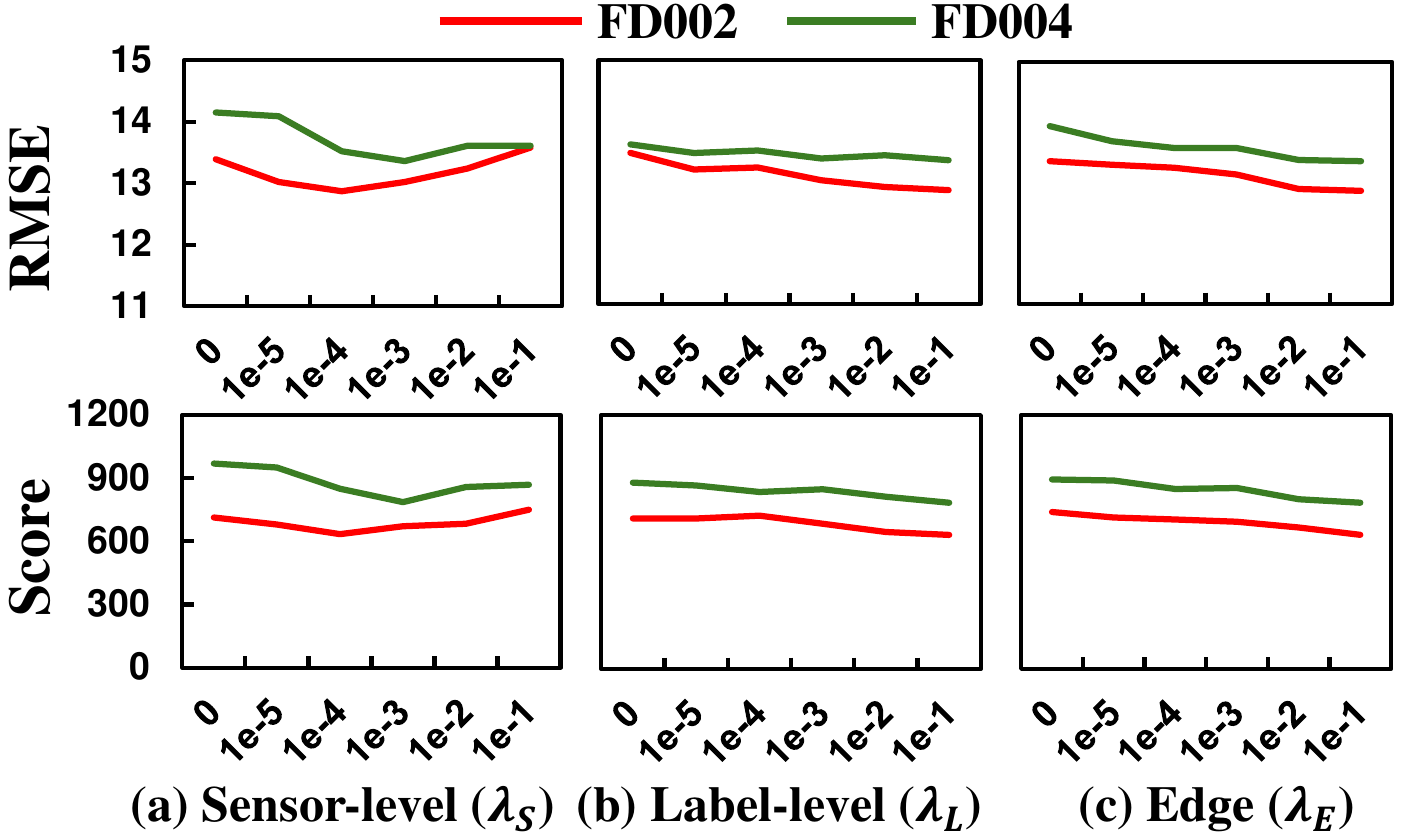}
    \caption{Sensitivity analysis for sensor-level, label-level, and edge alignment in regression tasks (Lower values of both indicators are better).}
    \label{fig:Analysis_overall_pre}
\end{figure}
\begin{figure}[htbp!]
    \centering\includegraphics[width = .9\linewidth]{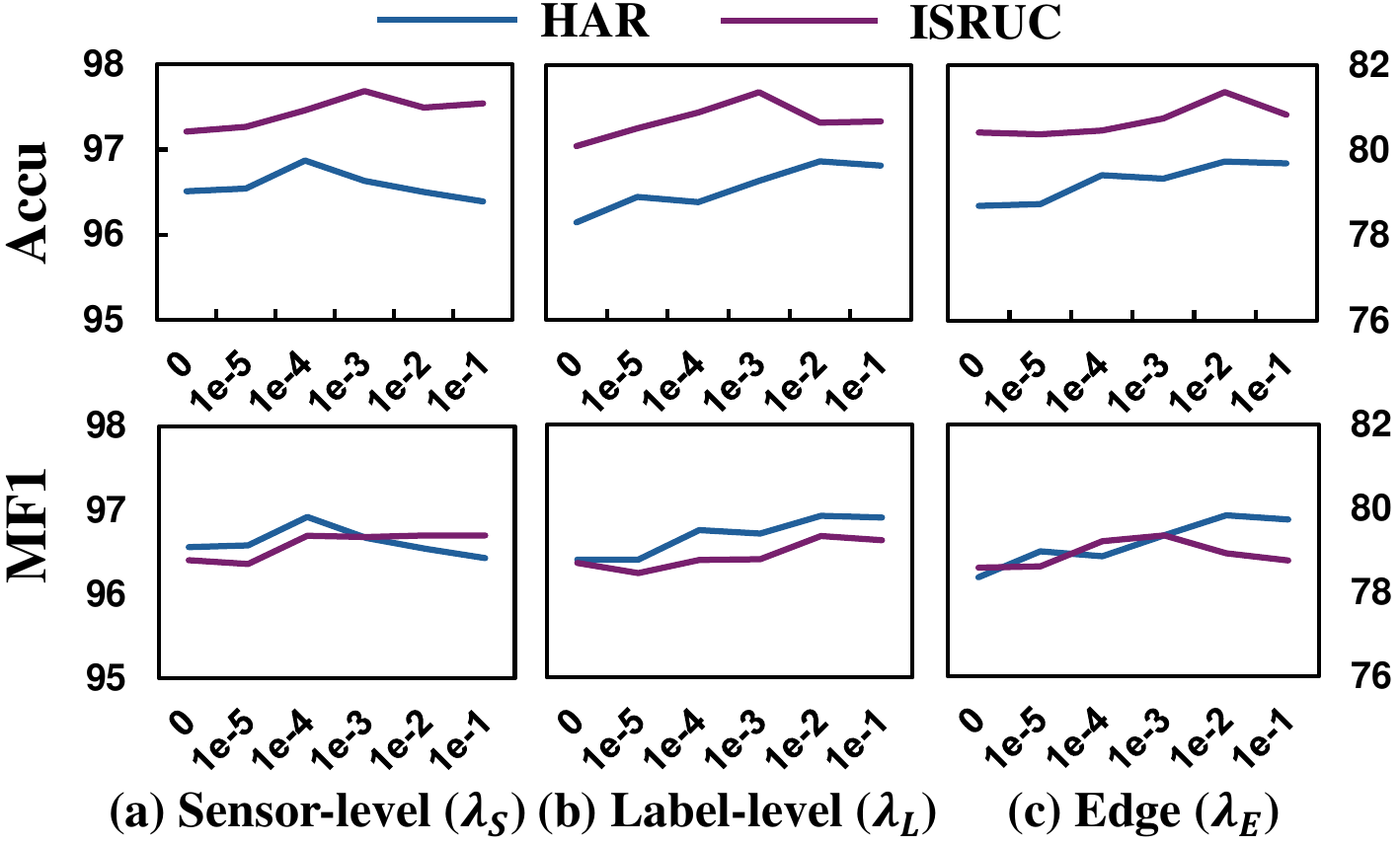}
    \caption{Sensitivity analysis for sensor-level, label-level, and edge alignment in classification tasks (Higher values of both indicators are better).}
    \label{fig:Analysis_overall_cls}
\end{figure}

Three hyperparameters, $\lambda_S$, $\lambda_L$, and $\lambda_E$, are introduced to adjust the contributions of sensor-level, label-level, and edge alignment, respectively. To analyze their effects, we evaluate various values within the range [0, 1e-5, 1e-4, 1e-3, 1e-2, 1e-1], where a value of 0 indicates the exclusion of the corresponding loss term. Figs \ref{fig:Analysis_overall_pre} and \ref{fig:Analysis_overall_cls} illustrate the impact of these hyperparameters on regression and classification tasks, respectively. For regression tasks, lower indicator values denote better performance, while for classification tasks, higher indicator values correspond to improved performance.

\textbf{Node Alignment Analysis - Sensor:}
This analysis focuses on the impact of sensor-level prompts, which convey universal knowledge about individual sensors. From the analysis $\lambda_S$ in both tasks of Figs \ref{fig:Analysis_overall_pre} and \ref{fig:Analysis_overall_cls}, we observe two key points. First, a small $\lambda_S$, such as $\text{1e-5}$, can achieve better performance than $\lambda_S = \text{0}$ (i.e., without sensor-level alignment), indicating the effectiveness of sensor-level alignment. Second, too small and large values fail to yield optimal performance. When the value is too small, the sensor-level alignment is insufficient to effectively transfer sensor knowledge into the MTS graph. For instance, in the case of FD002 and HAR, the performance with $\lambda_S = \text{1e-5}$ is worse than the results with $\lambda_S = \text{1e-4}$. In contrast, large values also result in sub-optimal performance, e.g., $\lambda_S = \text{1e-1}$. This occurs because sensor-level prompts contain universal knowledge about sensors but fail to capture sample-specific details. A large $\lambda_S$ results in strong sensor-level alignment, where universal semantic features dominate over sample-specific features, leading to an excessive loss of sample-specific information and diminished performance. In summary, sensor-level prompts contribute positively to performance, but the hyperparameter governing sensor-level alignment should not be too small or large. $\lambda_S$ = 1e-4 or 1e-3 can help achieve optimal performance.

\textbf{Node Alignment Analysis - Label:}
This analysis focuses on the effect of label-level prompts, which provide additional information for variations in different categories. As the analysis of $\lambda_L$ depicted in Figs \ref{fig:Analysis_overall_pre} and \ref{fig:Analysis_overall_cls}, we observe that increasing the value of $\lambda_L$ enhances performance, underscoring the efficacy of the label-level alignment. However, the performance diminishes with large values. For instance, setting $\lambda_L$ to 1e-1 on HAR and ISRUC fails to yield optimal results. For regressions, the trend stabilizes with $\lambda_L$ increasing to 1e-2. Notably, different from $\lambda_S$, $\lambda_L$ attains optimal solutions with relatively larger values, such as 1e-2 for HAR, while $\lambda_S$ achieves optimal performance with relatively smaller values, such as 1e-4 for HAR. For this reason, we divide the node alignment into sensor-level and label-level alignment, making it easier to achieve optimal performance.

\textbf{Edge Alignment Analysis:}
This analysis focuses on the effect of edge alignment, which transfers universal sensor relationships within the knowledge-link graph into sensor correlations learned from MTS data. From the analysis of $\lambda_E$ in Figs \ref{fig:Analysis_overall_pre} and \ref{fig:Analysis_overall_cls}, we observe that even a small value can contribute to better performance than the case of $\lambda_E=\text{0}$, indicating the effectiveness of edge alignment. However, when the value increases to 1e-3 or 1e-2, the performance shows no further improvement or even degrades. This suggests that large $\lambda_E$ might lead to the excessive loss of sample-specific sensor correlations. Therefore, 1e-3 or 1e-2 can help obtain optimal performance.
\subsection{Discussion for K-Link}
\label{sec:discuss}
\begin{figure*}[htbp!]
    \centering\includegraphics[width = .85\linewidth]{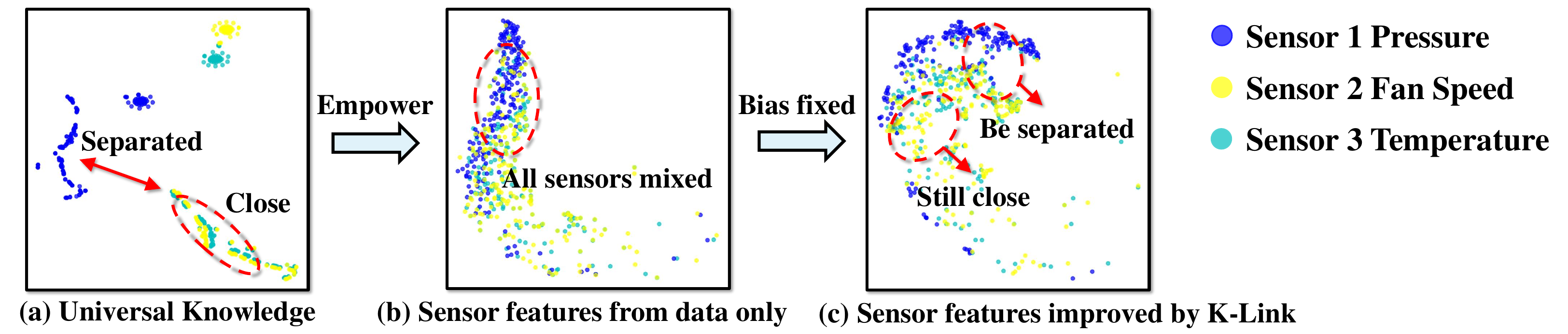}
    \caption{Visualizations to show that universal knowledge improves sensor feature distributions in MTS data by reducing biased correlations. (a) Universal knowledge shows sensors 2 and 3 are closely related, while sensor 1 is distant, reflecting their semantic meanings (e.g., fan speed correlates with temperature, not pressure). (b) Features learned without K-Link mix all sensors, indicating high correlations of them that contradict their semantic relationships, highlighting biases from small datasets. (c) K-Link integrates universal knowledge, refining distributions to separate sensor 1 from sensors 2 and 3 while preserving the close relationship between the latter, indicating enhanced sensor correlations that better reflect their semantic relationships.}
    \label{fig:visual}
\end{figure*}
\textbf{Effectiveness with visualization:}
To demonstrate K-Link's effectiveness in improving sensor features and their correlations for enhancing the graph generation process, we visualize and compare sensor features with and without K-Link. Specifically, we select features from three sensors and 100 samples from the FD002 dataset, using t-SNE for visualization. The results are shown in Fig. \ref{fig:visual}, where each point represents a sensor from a sample, and proximity between points indicates stronger correlations.

Fig. \ref{fig:visual} (a) presents universal sensor knowledge from the knowledge-link graph, i.e., sensor semantic features of prompts encoded by LLMs, with distances indicating their relationships. Here, sensors 2 and 3 are close, indicating a high correlation between them, while sensor 1 is distant, suggesting lower correlations with others. These relationships align with their semantic meanings, i.e., fan speed should be correlated with temperature instead of with pressure. However, the features solely learned from data without K-Link show different sensor distributions in Fig. \ref{fig:visual} (b), where all three sensors appear mixed, signifying high correlations among them. This result contradicts their semantic relationships, highlighting the bias introduced by small training datasets, which skews sensor correlations and adversely affects MTS graph generation.

K-Link addresses this issue by integrating universal knowledge from the knowledge-link graph to refine sensor feature distributions. As depicted in Fig. \ref{fig:visual} (c), the improved features separate sensor 1 from sensors 2 and 3, while maintaining the close relationship between the latter two. This adjustment indicates enhanced sensor correlations that better reflect their semantic relationships. These findings demonstrate that the knowledge-link graph effectively mitigates the biased correlations among sensors, offering a robust solution to improve MTS graph generation and advance representation learning with GNNs for MTS data.

\begin{table}[htbp]
  \centering
  \scriptsize
  \caption{K-Link with different GNN backbones for regression tasks}
  \begin{threeparttable}
    \begin{tabular}{lcccc}
    \toprule
    \toprule
    \multicolumn{1}{l}{\multirow{2}[2]{*}{Variants}} & \multicolumn{2}{c}{\textbf{FD002}} & \multicolumn{2}{c}{\textbf{FD004}} \\
          & RMSE \textbf{$\downarrow$}  & Score \textbf{$\downarrow$} & RMSE \textbf{$\downarrow$}  & Score \textbf{$\downarrow$} \\
    \midrule
    HierCorrPool & 13.23$\pm$0.31 & 709$\pm$61 & 13.86$\pm$0.32 & 854$\pm$68 \\
    K-Link (HierCorrPool) & 12.93$\pm$0.21 & 658$\pm$55 & 13.75$\pm$0.38 & 762$\pm$53 \\
    \midrule
    HAGCN & 14.92$\pm$0.12 & 1086$\pm$87 & 14.66$\pm$0.25 & 880$\pm$150 \\
    K-Link (HAGCN) & 13.81$\pm$0.19 & 885$\pm$43 & 14.19$\pm$0.28 & 852$\pm$120 \\
    \bottomrule
    \bottomrule
    \end{tabular}%
\begin{tablenotes}
        \item \textbf{$\downarrow$}: Lower values are better
      \end{tablenotes}
\end{threeparttable}
  \label{tab:backbones_pre}%
\end{table}%
\begin{table}[htbp]
  \centering
  \scriptsize
  \caption{K-Link with different GNN backbones for classification tasks}
  \begin{adjustbox}{width = 1\linewidth,center}
  \begin{threeparttable}
    \begin{tabular}{lcccc}
    \toprule
    \toprule
    \multicolumn{1}{l}{\multirow{2}[2]{*}{Variants}} & \multicolumn{2}{c}{\textbf{UCI-HAR}} & \multicolumn{2}{c}{\textbf{ISRUC}} \\
          & Acc. \textbf{$\uparrow$} & MF1 \textbf{$\uparrow$}   & Acc. \textbf{$\uparrow$} & MF1 \textbf{$\uparrow$} \\
    \midrule
    HierCorrPool & 93.81$\pm$0.26 & 93.79$\pm$0.28 & 79.31$\pm$0.60 & 76.25$\pm$0.72 \\
    K-Link (HierCorrPool) & 94.60$\pm$0.26 & 94.69$\pm$0.28 & 79.80$\pm$0.52 & 78.00$\pm$0.63 \\
    \midrule
    HAGCN & 80.79$\pm$0.77 & 81.08$\pm$0.75 & 66.59$\pm$0.29 & 60.20$\pm$2.24 \\
    K-Link (HAGCN) & 84.41$\pm$0.45 & 85.13$\pm$0.43 & 71.71$\pm$0.34 & 66.73$\pm$0.92 \\
    \bottomrule
    \bottomrule
    \end{tabular}%
\begin{tablenotes}
        \item \textbf{$\uparrow$}: Higher values are better
      \end{tablenotes}
\end{threeparttable}
    \end{adjustbox}
  \label{tab:backbones_cls}%
\end{table}%

\textbf{Effectiveness with different backbones:}
K-Link can seamlessly integrate into existing GNN-based frameworks to enhance their graph generation process. To validate its effectiveness, we incorporated K-Link into two GNN methods, HierCorrPool \cite{wang2023multivariate} and HAGCN \cite{LI2021107878}. As shown in Tables \ref{tab:backbones_pre} and \ref{tab:backbones_cls}, we observe significant performance improvements and reduced standard deviations. For example, HierCorrPool with K-Link improves by 7.19\% and 1.75\% on FD002 and ISRUC, respectively. HAGCN with K-Link improves by 18.51\% and 6.53\%, respectively. These improvements highlight the effectiveness of K-Link.

\textbf{Comparisons with Domain-specific GNN Approaches:}
We further compare K-Link with GNN methods that incorporate prior knowledge for graph generation enhancement. As these methods are domain-specific, comparisons are made only on their specific domains. In this section, we focus on FD004 and ISRUC tasks, as they are more challenging and complex, resembling real-world scenarios. For FD004, we evaluate against STFA \cite{kong2022spatio}, which leverages prior knowledge of predefined aero-engine component connections to assist in graph generation for sensors deployed in an aero-engine. For ISRUC, we compare with MSTGNN \cite{jia2021multi}, which utilizes two graphs: one constructed purely from data and the other based on the actual physical distances between brain regions.

As shown in Tables \ref{tab:prior_comp_pre} and \ref{tab:prior_comp_cls}, K-Link consistently outperforms these methods, highlighting two key points: (1) While existing works rely on domain-specific prior knowledge for graph enhancement, LLMs equip K-Link with greater capabilities to generate better graphs and further improve representation learning. (2) Unlike domain-specific approaches, K-Link's reliance on LLMs enables it to generalize across diverse domains, demonstrating its superior effectiveness and flexibility.

\begin{table}[htbp]
  \centering
  \caption{Comparison of GNN methods incorporating knowledge for graph enhancement in regression tasks.}
  \begin{threeparttable}
    \begin{tabular}{lcc}
    \toprule
    \toprule
    \multirow{2}[2]{*}{Models} & \multicolumn{2}{c}{\textbf{FD004}} \\
          & RMSE \textbf{$\downarrow$} & Score \textbf{$\downarrow$}\\
    \midrule
    STFA  & 15.06$\pm$0.18 & 1184$\pm$97 \\
    K-Link & \textbf{13.36$\pm$0.22} & \textbf{786$\pm$61} \\
    \bottomrule
    \bottomrule
    \end{tabular}%
\begin{tablenotes}
        \item \textbf{$\downarrow$}: Lower values are better
      \end{tablenotes}
\end{threeparttable}
  \label{tab:prior_comp_pre}%
\end{table}%
\begin{table}[htbp]
  \centering
  \caption{Comparison of GNN methods incorporating knowledge for graph enhancement in classification tasks.}
  \begin{threeparttable}
    \begin{tabular}{lcc}
    \toprule
    \toprule
    \multirow{2}[2]{*}{Models} & \multicolumn{2}{c}{\textbf{ISRUC}} \\
          & Acc. \textbf{$\uparrow$} & MF1 \textbf{$\uparrow$} \\
    \midrule
    MSTGNN & 80.48$\pm$0.31 & 78.53$\pm$0.64 \\
    K-Link & \textbf{81.37$\pm$0.20} & \textbf{79.36$\pm$0.49} \\
    \bottomrule
    \bottomrule
    \end{tabular}%
\begin{tablenotes}
        \item \textbf{$\uparrow$}: Higher values are better
      \end{tablenotes}
\end{threeparttable}
  \label{tab:prior_comp_cls}%
\end{table}%

\textbf{Discussion of Complexity:}
In this section, we analyze the complexity of K-Link, focusing on both trainable parameters and time requirements. While LLMs inherently have numerous parameters that could increase model complexity, K-Link mitigates this by leveraging a pretrained text encoder from the LLM, which remains fixed and does not introduce additional trainable parameters. 

Regarding time complexity, the training time increases slightly due to the need to extract the knowledge-link graph from the LLM using prompts. However, this increase remains manageable because the training datasets are relatively small, and the fixed LLM text encoder minimizes additional computational overhead. To evaluate training efficiency, we integrated K-Link with different LLMs, including GPT-2, Llama 3.2 1B, and DeepSeek R1 1.5B, and measured their training times across multiple datasets. (Performance comparisons with these LLMs are discussed in the next section.) The training times for different LLMs across multiple datasets are summarized in Table \ref{tab:trainingtime}. For instance, training K-Link with GPT-2 on FD002, which contains 31,816 training samples, requires only 0.17 hours. Even with larger models such as Llama and DeepSeek, the training times are 0.69 hours and 0.92 hours, respectively, which remain manageable. During inference, the LLM is removed entirely, eliminating any additional computational burden and enabling rapid predictions. For instance, a single sample can be processed in just 0.01 seconds. These results indicate that K-Link strikes an effective balance between performance and efficiency, making it a practical and scalable solution for real-world applications.
\begin{table}[htbp]
\scriptsize
  \centering
  \caption{Training time analysis (hour)}
    \begin{tabular}{lcccc}
    \toprule
    \toprule
    Model & \textbf{FD002} & \textbf{FD004} & \textbf{HAR} & \textbf{ISRUC} \\
    \midrule
    K-Link with GPT-2 & 0.17  & 0.22  & 0.05  & 0.05 \\
    K-Link with Llama & 0.69  & 0.75  & 0.17  & 0.27 \\
    K-Link with DeepSeek & 0.92 & 0.97 & 0.22 & 0.62\\
    \bottomrule
    \bottomrule
    \end{tabular}%
  \label{tab:trainingtime}%
\end{table}%

\textbf{Discussion of Leveraging Different LLMs:}
Due to its exceptional efficiency in encapsulating rich universal knowledge, GPT-2 has been utilized in this work for experimentation. However, K-Link can also be combined with more advanced LLMs to potentially achieve further improvements. As K-Link relies on a pretrained text encoder to extract semantic features from prompts, only open-source LLMs can be used. As a result, we adopted two open-source LLMs—Llama 3.2 1B and DeepSeek R1 1.5B—rather than well-known GPT models, which are closed-source.

Tables \ref{tab:LLM_pre} and \ref{tab:LLM_cls} present the results of K-Link with different LLMs for regression and classification tasks, respectively. From the results, we observe that using advanced LLMs to generate the knowledge-link graph improves performance in some cases. For instance, variants with Llama and DeepSeek show improvements of 3.78\% and 2.83\%, respectively, compared to K-Link with GPT-2 for the Score value on FD002, even though K-Link with GPT-2 already achieves significant improvements. When compared to the model without K-Link, the improvements are as high as 24.12\% and 23.38\%, respectively. This performance boost is due to the larger LLMs bringing more comprehensive universal knowledge of sensors and their relationships, enabling the extracted knowledge-link graph to more accurately represent this information. With a more accurate knowledge-link graph, the graph generation process is further enhanced, leading to the generation of more effective MTS graphs and improving representation learning with GNNs.

However, although advanced LLMs yield better performance, they also incur longer training times. As seen in Table \ref{tab:trainingtime}, the training time increases from 0.22h with GPT-2 to 0.69h with Llama, and further to 0.92h with DeepSeek. While K-Link does not impact the inference stage, it does increase the training burden, requiring more resources for training with larger LLMs. Therefore, we recommend GPT-2 as a suitable choice, as it delivers solid performance while keeping training resource requirements manageable.

\begin{table}[htbp]
  \centering
  \caption{The effects of different LLMs on K-Link for regression tasks.}
    \begin{threeparttable}
    \begin{tabular}{lcccc}
    \toprule
    \toprule
    \multicolumn{1}{l}{\multirow{2}[2]{*}{Variants}} & \multicolumn{2}{c}{\textbf{FD002}} & \multicolumn{2}{c}{\textbf{FD004}} \\
          & RMSE \textbf{$\downarrow$}  & Score \textbf{$\downarrow$} & RMSE \textbf{$\downarrow$}  & Score \textbf{$\downarrow$} \\
    \midrule
    w/o K-Link & 13.85$\pm$0.19 & 804$\pm$91 & 14.39$\pm$0.13 & 968$\pm$56 \\
        \midrule
    K-Link w/ GPT-2 & 12.87$\pm$0.14 & 634$\pm$30 & 13.36$\pm$0.22 & 786$\pm$61 \\
    K-Link w/ Llama & 12.72$\pm$0.14 & 610$\pm$29 & 13.18$\pm$0.25 & 712$\pm$36 \\
    K-Link w/ DeepSeek & 12.62$\pm$0.16 & 616$\pm$34 &12.85$\pm$0.13  & 754$\pm$45\\
    \bottomrule
    \bottomrule
    \end{tabular}%
\begin{tablenotes}
        \item \textbf{$\downarrow$}: Lower values are better
      \end{tablenotes}
\end{threeparttable}
  \label{tab:LLM_pre}%
\end{table}%

\begin{table}[htbp]
  \centering
  \caption{The effects of different LLMs on K-Link for classification tasks.}
    \begin{adjustbox}{width = 1\linewidth,center}
  \begin{threeparttable}
    \begin{tabular}{lcccc}
    \toprule
    \toprule
    \multicolumn{1}{l}{\multirow{2}[2]{*}{Variants}} & \multicolumn{2}{c}{\textbf{UCI-HAR}} & \multicolumn{2}{c}{\textbf{ISRUC}} \\
          & Acc. \textbf{$\uparrow$} & MF1 \textbf{$\uparrow$}   & Acc. \textbf{$\uparrow$} & MF1 \textbf{$\uparrow$} \\
    \midrule
    w/o K-Link & 95.62$\pm$0.16 & 95.64$\pm$0.18 & 79.83$\pm$0.14 & 78.11$\pm$0.55 \\
        \midrule
    K-Link w/ GPT-2 & 96.87$\pm$0.12 & 96.92$\pm$0.12 & 81.37$\pm$0.20 & 79.36$\pm$0.49 \\
    K-Link w/ Llama & 97.15$\pm$0.16 & 97.18$\pm$0.14 & 81.46$\pm$0.26 & 78.92$\pm$0.49 \\
    K-Link w/ DeepSeek & 97.22$\pm$0.14 & 97.22$\pm$0.16 &81.65$\pm$0.29  & 79.65$\pm$0.50\\
    \bottomrule
    \bottomrule
    \end{tabular}%
\begin{tablenotes}
        \item \textbf{$\uparrow$}: Higher values are better
      \end{tablenotes}
\end{threeparttable}
    \end{adjustbox}
  \label{tab:LLM_cls}%
\end{table}%





\section{Conclusion}
When adapting GNNs to MTS data, existing methods for MTS graph generation heavily rely on data and are thus vulnerable to biases from small training datasets, hindering effective MTS representation learning with GNNs.
To address this challenge, we propose a novel framework, K-Link, leveraging universal knowledge embedded within LLMs to reduce the biases for powered MTS graph generation. First, we extract a Knowledge-Link graph from LLMs, capturing universal sensor knowledge and the linkage of the knowledge. Second, we propose a graph alignment module to empower MTS graph generation with the knowledge-link graph.
This module facilitates the transfer of universal knowledge within the knowledge-link graph to the MTS graph. By doing so, we can improve the graph quality, ensuring effective representation learning with GNNs for MTS data. Extensive experiments demonstrate the efficacy of K-Link for superior performance across various MTS downstream tasks.

\bibliography{refer}
\bibliographystyle{IEEEtran}

\end{document}